%% file: main.tex
\documentclass[conference]{IEEEtran}
\usepackage{times}

\usepackage[numbers]{natbib}
\usepackage{multicol}
\usepackage[bookmarks=true]{hyperref}
\usepackage{amssymb,amsmath}
\usepackage{gensymb}
\usepackage{cite} % Compacts [1]--[4] citations and adds spaces
\usepackage{balance}
\usepackage{booktabs}
\usepackage{caption}
\usepackage[font=footnotesize,labelformat=simple]{subcaption}
\usepackage[font=footnotesize]{caption}
\usepackage{graphicx}
\usepackage{tabularx}
\usepackage{pifont}% http://ctan.org/pkg/pifont
\newcommand{\std}[1]{\tiny{$\pm$ #1}}

\usepackage{dsfont}

\graphicspath{{}}
\let\biblio\bibliography

\renewcommand{\bibliography}[1]{\expandafter\biblio{#1}}
\begin{document}

% paper title
\title{Demonstrating HOUND: A Low-cost Research Platform for High-speed Off-road Underactuated Nonholonomic Driving}

% You will get a Paper-ID when submitting a pdf file to the conference system
\author{Sidharth Talia, Matt Schmittle, Alexander Lambert, Alexander Spitzer, Christoforos Mavrogiannis, \\
Siddhartha S. Srinivasa \vspace{-10pt}}

\maketitle
\begin{abstract}
Off-road autonomy, crucial for applications such as search-and-rescue, agriculture, and planetary exploration, poses unique problems due to challenging terrains, as well as due to the risk involved in testing or deploying such systems. 
Accessible platforms have the potential to widen the field to a broader set of researchers and students.
Existing efforts in making on-road autonomy more accessible have seen success, yet aggressive off-road autonomy remains underserved.
We seek to fill this gap by introducing HOUND, a 1/10th-scale, inexpensive, off-road autonomous car platform that can handle challenging outdoor terrains at high speeds.
To aid development speed, we integrate HOUND with BeamNG, a state-of-the-art driving simulator to enable both software in the loop as well as hardware in the loop testing.
To reduce the extent of ruggedization required, and thus cost, we integrate a rollover prevention system as a safety feature into the platform.
Real-world trials over 50 kilometers demonstrate the platform's longevity and effectiveness over varied terrains and speeds. Build instructions, datasets and code disseminated via: \url{https://sites.google.com/view/prl-hound/home}
\end{abstract}

\IEEEpeerreviewmaketitle

\input{sections/introduction}
\begin{figure*}
\vspace{5pt}
    \includegraphics[width=\linewidth]{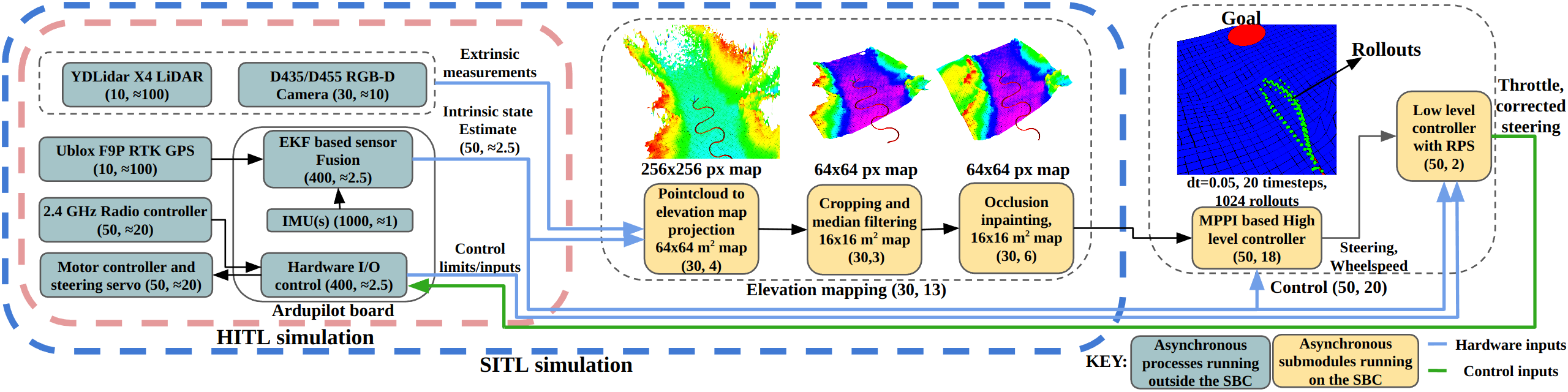}
    \caption{
    Autonomy stack components, specified with (rate in Hz, latency in ms), with $\approx$ implying estimated latency based on maximum update rate.
    SITL spoofs perception, useful for isolating control problems, whereas HITL spoofs sensors, useful for testing deployed software.
    }
    \label{fig:architecture}
\vspace{-15pt}
\end{figure*}
\input{sections/literature}
\input{sections/system_architecture}
\input{sections/Evaluation}
\input{sections/Discussion}
\input{sections/Conclusion}

\input{sections/Acknowledgments}
\bibliographystyle{plainnat}
\bibliography{references}
\input{sections/Appendix}

\end{document}

%% file: sections/introduction.tex
\section{Introduction~}
\label{sec:intro}
Off-road autonomy has various applications, including search-and-rescue~\citep{ISR}, agriculture~\citep{agriculture}, and planetary exploration~\citep{planetary}. 
Here, off-road environments refer to environments with highly uneven terrain, with structures such as hills or ditches as tall or deep as, if not taller or deeper than, the vehicle itself.
Off-road environments are not tailored to be traversed by vehicles; they present low traction and bumpy surfaces leading to difficulties in state estimation as well as control.
Such environments can also be physically unforgiving, resulting in potential sensor degradation if not outright damage.
Unlike on-road driving, there are no rules that explicitly dictate expected vehicle behavior. This can make benchmarking and comparison of off-road systems non-trivial.

Much of the progress in modern robotics has come from large efforts toward improving infrastructure for research and education. On-road autonomous driving, a generally resource-intensive field, is being democratized by efforts such as MuSHR~\citep{srinivasa2019mushr} and F1-tenth~\citep{f1tenth}, which lowered the barriers of entry by providing a small, low-cost, easy-to-use hardware/software platform.
The AutoRally~\citep{goldfainthesis} platform had a similar aim for off-road aggressive driving, being less expensive and smaller than the platforms used in other works~\citep{speedrisk, terrainnet}. 
While it created a significant cost reduction, this work seeks to further lower the cost through new low-cost hardware and novel algorithms previously not available, to enable off-road autonomy research for a larger audience.

As there are no explicit rules in off-road autonomy, evaluating an autonomy stack requires knowing whether its actions would incur damage or not.
As such, we provide integration with BeamNG~\citep{beamng_tech},  a high-fidelity vehicle simulator used both in the industry as well as academia~\citep{BeamNG1, BeamNG2}, known particularly for its accurate crash simulation. 

A major concern for high-speed off-road navigation tends to be rollovers ~\citep{goldfainthesis}, as repeated rollovers can damage the components over time, costing both money and time in repairs. In essence, the vehicle rolls over due to excess lateral acceleration.
As such, we integrate a low-level rollover prevention system (\textbf{RPS}) as a safety feature into our platform that allows researchers to push the limits without needing to repair the platform regularly or using expensive, ruggedized hardware.

\begin{figure}[!t]
\centering
\includegraphics[width=\linewidth]{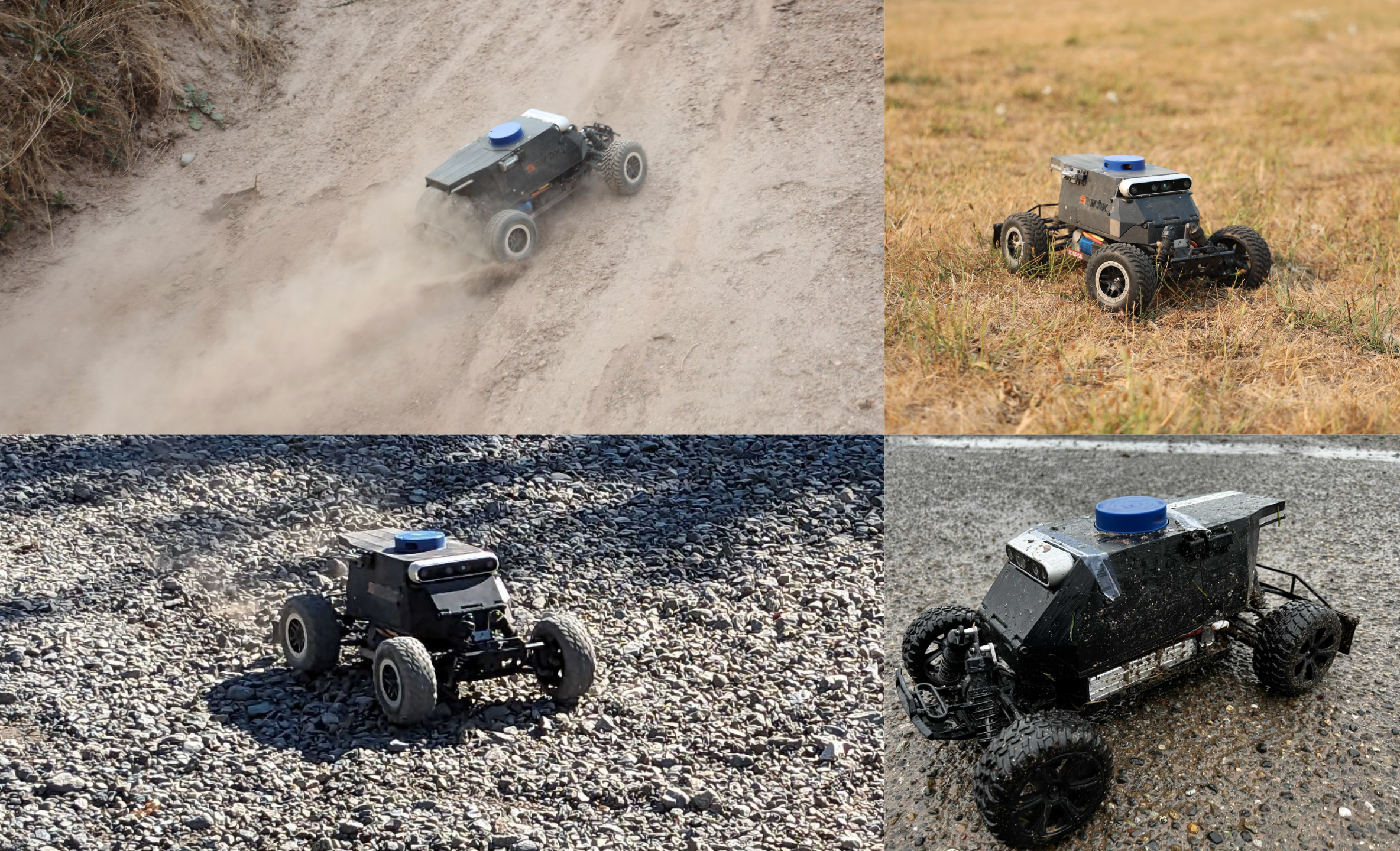}
\caption{HOUND has been tested on dirt hills, grasslands, gravel trails, and tarmac.}
\label{fig:best}
\vspace{-20pt}
\end{figure}

The main contribution of our work is an integrated platform (see Fig\ref{fig:best}), with open-source software, and hardware, that can be used by off-road autonomy researchers, and robotics researchers in general, at a relatively low cost of $\approx$\$3000.
Our open-source software stack removes a major barrier to research: the need to engineer an entire off-road autonomy stack before starting focused research. Further, we believe that a common accessible platform enables easy data sharing across research projects.
We demonstrate the utility of the simulator by using it to validate the RPS. We evaluate the RPS both in an isolated fashion, as well as in conjunction with a high-level controller to show that it does not interfere excessively.
Finally, we test the longevity of this system in the real world for \textbf{50 km} over 4 different terrains, reaching speeds up to \textbf{7 $m/s$} and lateral accelerations up to \textbf{9 $m/s^2$}.

%% file: sections/literature.tex
\section{Related Work~}
\label{sec:related}
\subsection{Aggressive off-road autonomy}
Aggressive off-road autonomy presents unique challenges as compared to on-road driving. The surface-tire interactions become more relevant due to slippage, vibrations due to terrain unevenness can make inertial state estimation harder, and occlusion due to environmental structures such as trees and ditches can make accurate mapping challenging. As such, work has been done towards perceiving outdoor environments as probabilistic elevation maps~\citep{extended_elevation, gpu_elevation_mapping}, and their extensions~\citep{elevation_inpainting, terrainnet} which use data-driven methods to address environment occlusion via inpainting. Work by~\citet{non_planar_anal} has extended planar dynamics models \citep{marty_drift} towards non-planar surfaces, whereas other works~\citep{terrainCNN, multistep_dynamics} have presented approaches for learning such 3-dimensional dynamics.
In this work, we incorporate some of the recent advances in off-road autonomy for extrinsic state estimation~\citep{gpu_elevation_mapping, elevation_inpainting} and control~\citep{smooth_mppi, alternate_tire_model_evidence} into a real-world system.

\subsection{Research platforms}
Full-scale autonomous driving systems are expensive, require vast resources, such as a large testing area, and can pose safety risks. As such, small-scale research platforms have become a popular entry point to autonomous driving research~\citep{mit_racecar,one_off_1, one_off_2,srinivasa2019mushr,f1tenth}. Notably, the MuSHR\citep{srinivasa2019mushr} and F1-tenth\citep{f1tenth} platforms, inspired by the MIT-RACECAR\citep{mit_racecar}, are popular platforms that have enabled a wide variety of research, even in external institutions~\citep{jain2020bayesrace, pact, talia2023pushr}. The AutoRally~\citep{goldfainthesis} platform further enabled research in outdoor environments~\citep{information_theoretic, pandeep},
but is significantly more expensive due to the use of industrial-grade sensors and the ruggedization required for off-road driving (see Table~\ref{comparison_table}).
In this work, we incorporate a rollover prevention system into the autonomy stack to reduce the cost associated with ruggedization. Our work is related to the VertiWheeler~\citep{vertiwheeler}, a recent work focusing on off-road navigation for rock crawling at lower speeds, but our focus is more on high-speed off-road navigation with rollover prevention. FastRLap~\citep{fastrlap} is a recent work focusing on using reinforcement learning for outdoor navigation with a small-scale platform, whereas we focus on creating a standardized small-scale research and education platform for outdoor autonomy.

\subsection{Rollover prevention}\label{RPS_related}
Rollover prevention is a popular problem in the domain of utility-grade automobile research, as it poses a safety risk to large vehicles or vehicles with a high center of mass during turning or side-slope traversal~\citep{rollover_prevention_1, rollover_prevention_3, rollover_prevention_4}. 
In this context, an Advanced Driver Assistance System(ADAS) uses a rollover index \textbf{(RI)} such as Load Transfer ratio~\citep{load_transfer_ratio, load_transfer_ratio_2}, Force angle measure~\citep{force_angle_1, force_angle_2000} or Time to Rollover~\citep{TTR_OG} to detect an imminent rollover and either alert the driver or additionally apply remedial controls~\citep{rollover_ADAS}.
For autonomous systems, works such as~\citep{rollover_MPC, terrainCNN} propose using the rollover indicator as part of the cost for the model predictive controller (MPC). 
For safety-critical tasks such as rollover prevention, erroneous extrinsic/intrinsic state estimation or dynamics mismatch can result in catastrophic failure. 
In our work, we incorporate a low-level reactive controller that uses Force angle measure~\citep{force_angle_2000} as the RI, inspired by \citet{rollover_LQR}, into the autonomy stack that only depends on an inertial measurement unit \textbf{(IMU)} and a wheel speed sensor. Additionally, we show that such a system can work in conjunction with an MPC that uses the rollover cost without loss of performance while reducing the chance of incidental rollovers.

%% file: sections/system_architecture.tex
\begin{table*}[htb!]
    \vspace{5pt}
    \centering
    \setlength{\tabcolsep}{4pt} % Reduce space between columns
    \begin{tabular}{llllll}
    \toprule
        Platform  & Focus & Sensing & Full cost & Autonomy cost(sensing \& compute) & Hardware / Software Rollover protection \\ \midrule
        F1Tenth~\citep{f1tenth} & Indoor & IMU, RGBD, LiDAR & \$3,000 & \$2,000  & None / None \\
        MuSHR~\citep{srinivasa2019mushr} & Indoor & IMU, RGBD,  LiDAR & \$1,000 & \$500  & None / None \\
        AutoRally~\citep{goldfainthesis} & Outdoor & IMU, Stereo RGB, GPS  & \$15,000 & \$9,000 & Steel body, industrial grade sensors / None \\
        \textbf{HOUND} & \textbf{Outdoor} & \textbf{IMU, RGBD, LiDAR, GPS} & \textbf{\$3,000} & \textbf{\$2,000}  & \textbf{Enclosed plastic shell / RPS} \\ 
        \bottomrule
    \end{tabular}
    \caption{
    Cost/features comparison among platforms. Costs approximated to the nearest 1000.
    Autorally's use of industrial grade sensors results in an autonomy cost $\approx4$x that of the HOUND, which uses commercial grade sensors and RPS to protect against rollover damage.
    }
\label{comparison_table}
\vspace{-15pt}
\end{table*}
\section{System Overview~}
\label{sec:system}
\subsection{Hardware specifications}\label{hardware}
The HOUND's physical platform is built using the Blackout SC-1/10 platform, also used in the MuSHR~\citep{srinivasa2019mushr}
due to its low cost and low weight as compared to a 1/5th scale platform.
As the autonomy stack is parameterized by vehicle properties, the use of a bigger/different chassis is not precluded by our work.
Details regarding the sensor stack are shown in Fig. \ref{fig:architecture}, Table \ref{comparison_table}.
The onboard single-board-computer \textbf{(SBC)} is an NVIDIA Jetson Orin NX as it provides a good compromise between cost and size-weight-and-power (\textbf{SWaP}) when compared to the Nano and AGX variants.
The physical platform weighs close to 4 Kg and can reach speeds beyond 12 m/s on tarmac. For system hardware specifications, please see table \ref{system_spec_table}, Fig \ref{fig:architecture}.
In contrast to previous outdoor autonomy platforms such as AutoRally~\citep{goldfainthesis}, and FastRLap~\citep{fastrlap}, which perform GPS-IMU fusion on the SBC, we use the Ardupilot~\citep{ardupilot} framework.
Here, a separate microcontroller board running the Ardupilot software~\citep{pixhawk} runs an EKF~\citep{ardu_ekf}, currently for GPS-IMU fusion, but optionally for fusing additional odometry measurements, either from visual odometry or motion capture.
Ardupilot has been used in this capacity before by works~\citep{arduboat, ardupilot1}.
Additionally, it provides a hardware bridge between the SBC and the chassis with several fail safes as well as various tools such as Missionplanner~\citep{mission_planner} for convenient field testing. 
The above components are housed inside a 3D-printed body (see Fig.~\ref{fig:coordinate_frame}) designed to protect against environmental factors and incidental rollovers, up to a limit.
For more details, refer to Appendix \ref{ardupilot_appendix} and \ref{hardware_appendix}.

\begin{table}[t!]
    \centering
    \setlength{\tabcolsep}{1pt}
    \begin{tabular}{ll}
    \toprule
        Specification & Value \\
        \midrule
        Max. wheelspeed (max $V_w$) & $23.0m/s$ (no load)\\
        Static rollover limit ($RI_L$) & $\approx0.9$ \\
        Battery backup (idle/driving) & 80 mins/15 mins @ 5m/s\\
        Weight (with battery) & $\approx4.0Kg$\\
        \midrule
        Component & Name \\
        \midrule
        High-level computation & Nvidia Jetson Orin NX 16 GB\\
        Ardupilot board & mRo PixRacer Pro (3 IMUs, 1 Barometer) \\
        GPS & Ublox F9P-01B RTK\\
        RGB-D camera & Intel Realsense D435/D455\\
        LiDAR & YDLidar X4\\
        Motor control/feedback & Flipsky FSESC 4.12\\
        Chassis & Redcat racing Blackout SC-1/10\\
        Steering Servo & ProTek RC 170SBL \\
        Battery & Spektrum SPMX324S100 (47.36Wh) \\
        \bottomrule
    \end{tabular}
    \caption{Note that while the maximum wheelspeed can reach 23 $m/s$, autonomy experiments are not expected to exceed $7-9$ $m/s$.}
    \label{system_spec_table}
    \vspace{-20pt}
\end{table}

\subsection{State estimation}\label{state_est}
The intrinsic state estimation, which includes pose-twist estimation, and IMU filtering, is done on the Ardupilot board and communicated to the SBC (see Fig. \ref{fig:architecture} for details).
For extrinsic state estimation, we use GPU-based probabilistic elevation mapping~\citep{gpu_elevation_mapping} in conjunction with a learning-based elevation inpainting system~\citep{elevation_inpainting} to address occlusions.
The elevation map is body-centric. 
On the Orin NX, the latency of projecting the pointcloud to the elevation map is invariant up to a relatively large map size, however, the latency of inpainting scales quadratically with the map size.
Thus, we maintain a large ``raw'' elevation map, and inpaint only a smaller, cropped portion of it, big enough for navigation at 6-8 m/s with planning horizons up to 1 second.
Before inpainting, spikes in the elevation map that occur due to noise in depth sensing are median filtered.
This minimizes the latency without excessively compromising on map quality and makes inpainting easier over time when the car operates in a small loop.
Details regarding the map size, latencies and update rates can be found in Fig. \ref{fig:architecture}.

\subsection{High-level controller}\label{HLC}
The controller's state vector consists of world frame position $X^w, Y^w, Z^w$, world frame roll-pitch-yaw $\phi, \theta, \psi$, body frame velocity $V^b$, body frame linear acceleration $A^b$, and body frame rotation rates $\omega^b$ in the reference frame shown in Fig. \ref{fig:coordinate_frame}. 
The controller also takes the elevation map,
and a path $G = (g^1,\dots,g^m), g^j\in \mathbb{R}^2$ in the world frame. 
The controller produces wheel speed $V_w$ and steering angle $\delta$ as the output.
The end users may implement any controller that follows the aforementioned input-output scheme.
As the default high-level controller, we implement a variant of MPPI~\citep{smooth_mppi} as our sampling-based controller using Pytorch and PyCUDA.

\subsubsection{Dynamics model}\label{dynamics_model}
\begin{figure}[!t]
\centering
\includegraphics[width=0.9\linewidth]{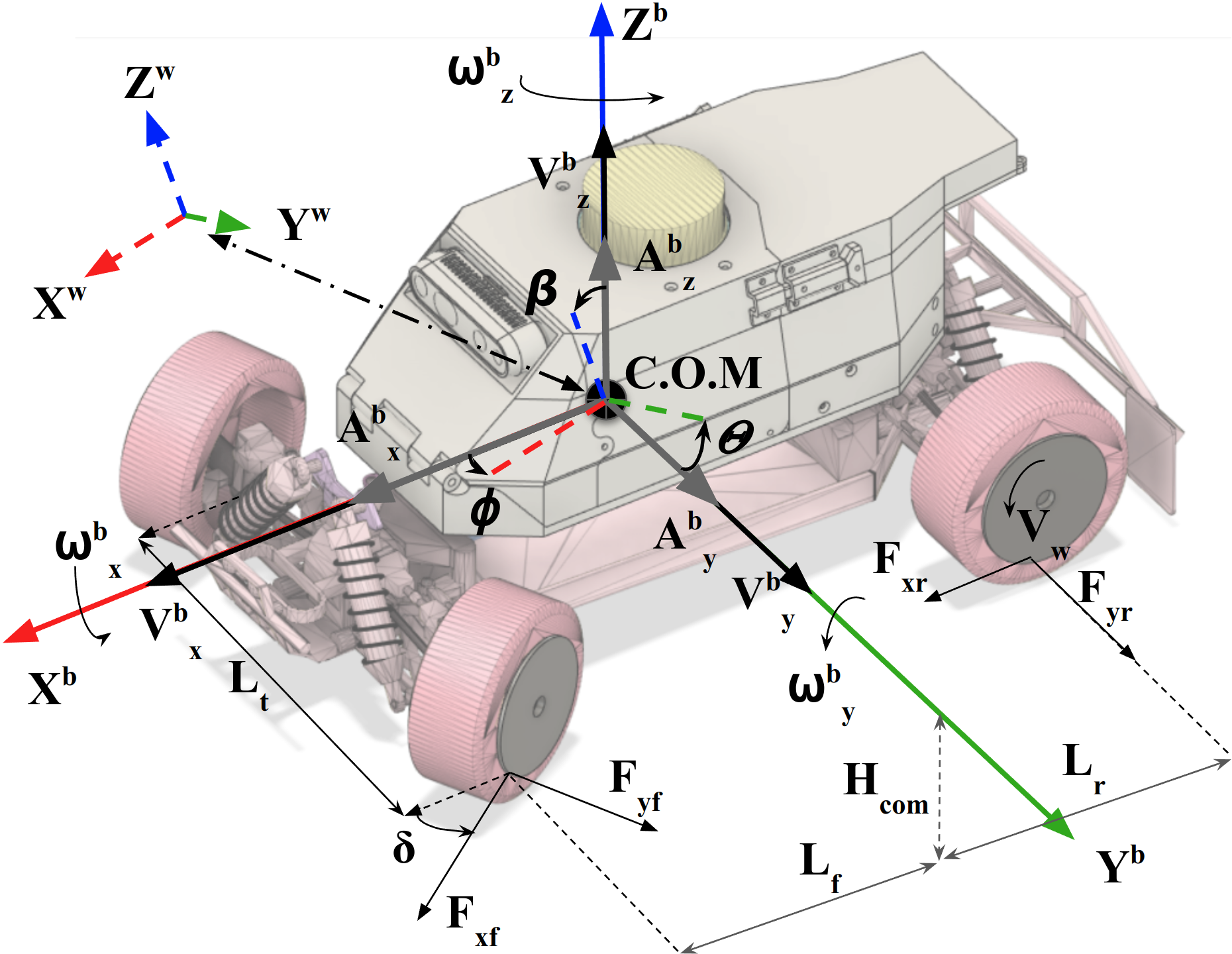}
\caption{Illustration of the coordinate frame used by HOUND}
\label{fig:coordinate_frame}
\vspace{-15pt}
\end{figure}

We use a single-track bicycle model~\citep{alternate_tire_model_evidence}, which considers non-planar surfaces, with a simplified Pacejka tire model~\citep{simplified_pacejka}, as it performs better than the kinematic no-slip model~\citep{kinematic_flat} near the tire's grip limit~\citep{dynamic_gerdes}.
The dynamics model is used to predict future states for a given control sequence $U = (u^1,\dots,u^m), u^j=(\delta^j, V_w^j)$.
We assume that the tires always remain in contact, as such the height and tilt --$z, \phi, \theta$--are obtained by projecting the location of the tires on the elevation map and  $V^b_z = 0$, as done by~\citet{terrainnet}.
The body rates $\omega^b_x, \omega^b_y$ are obtained by transforming world frame roll-pitch-yaw rates into the body frame~\citep{cost_function_arg}.
The body frame forces $F^b_x, F^b_y, F^b_z$ and rotational acceleration $\dot{\omega^b_z}$ are given by:
\begin{align}\label{wheel_vel}
\begin{split}
F^b_x &= F_{xr} + F_{xf} \cos(\delta) - F_{yf} \sin(\delta) + m g \sin{\theta} ,\\
F^b_y &= F_{yr} + F_{yf} \cos(\delta) + F_{xf} \sin(\delta) + m g \sin{\phi} ,\\
F^b_z &= m (g \cos{\beta} - V_x \omega_y + V_y \omega_x),\\
\dot{\omega^b_z} &= ((F_{xf} \sin(\delta) + F_{yf} \cos(\delta))L_f - F_{yr} L_r)/J_z
\end{split}
\end{align}
Where $F_{xf}$, $F_{yf}$ represent the wheel-frame longitudinal and lateral forces generated by the front tire, $F_{xr}$, $F_{yr}$ represent the same for the rear tires, $\beta$ represents the angle made by the body-frame Z-axis with the world's Z axis(see Fig.~\ref{fig:coordinate_frame}), $L_f, L_r$ represent the distance from the center of mass to the front and the rear axle respectively, $m$ represents the mass of the vehicle, $J_z$ represents the mass moment of inertia around its Z axis. 
These lateral and longitudinal accelerations are then used to update the body frame velocity $V^b$, which is transformed into the world frame to update the position.

\subsubsection{Cost function}\label{cost_function_section}
For the cost along the trajectory, following recent work~\citep{terrainnet, terrainCNN}, we aim to regulate vertical force $F^b_z$, angle to the world frame Z axis $\beta$, body frame forward velocity $V^b_x$, rollover index $RI$ (see Eq. \ref{RI_def}), cross-track error $E_x(p)$ and Euclidean distance to goal along the path $E_g(p)$, $p=(X^w, Y^w), p\in \mathbb{R}^2$.
\begin{align}\label{cost_function}
\begin{split}
C &= W_f T(F^b_z, F^b_L) +  W_\beta T(\beta, \beta_L) + W_v T(V^b_x, V_L) + \\ & W_r T(RI, RI_L) + W_s E_x(p) + W_g E_g(p)
\end{split}
\end{align}
Where $C$ is the total cost that gets summed along the trajectory. Note that the terms with a ``L'' subscript represent the limiting value for the corresponding variable, and $T(x,L)=\max(0,x-L)$.
$W_f, W_\beta, W_v, W_r, W_s, W_g$ are weights on these costs. 

\begin{figure*}
\vspace{5pt}
\begin{subfigure}{0.33\linewidth}
  \centering
  \includegraphics[width=\linewidth]{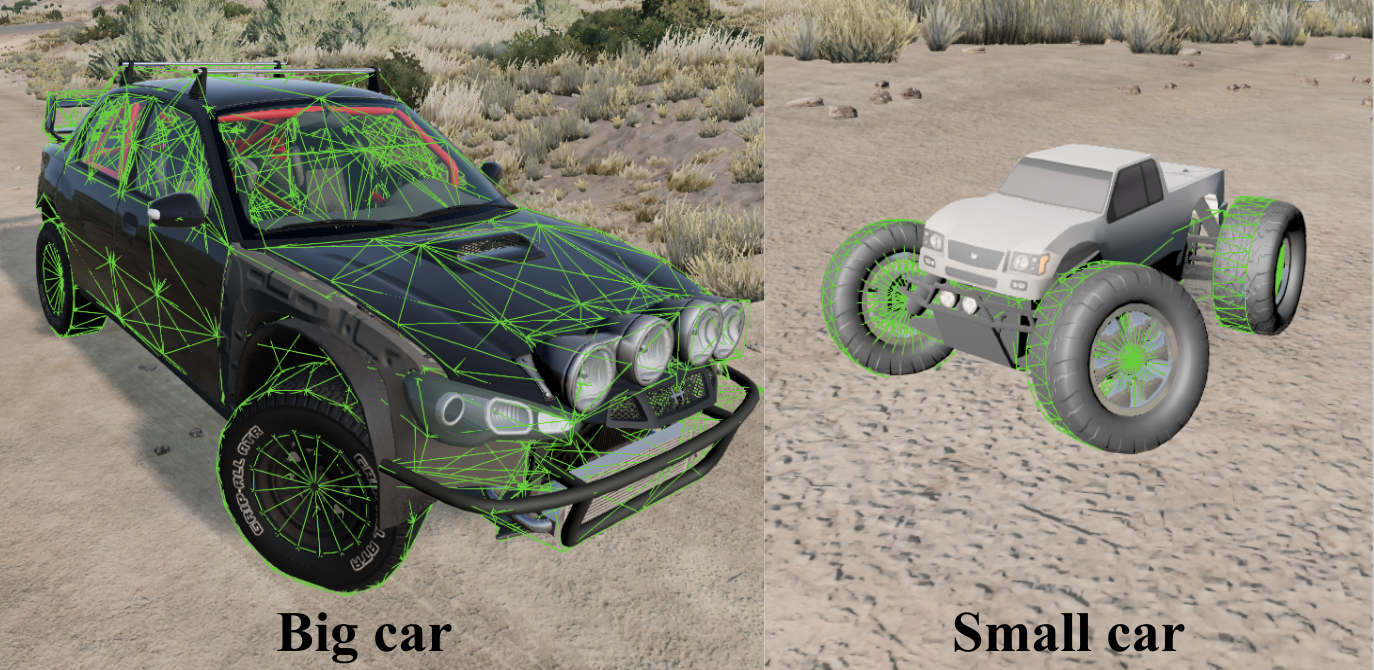}
  \caption{Beam-node based physics simulation}
  \label{fig:beamNG_sim}
\end{subfigure}
\hspace{0.1pt}
\begin{subfigure}{0.295\linewidth}
  \centering
  \includegraphics[width=\linewidth]{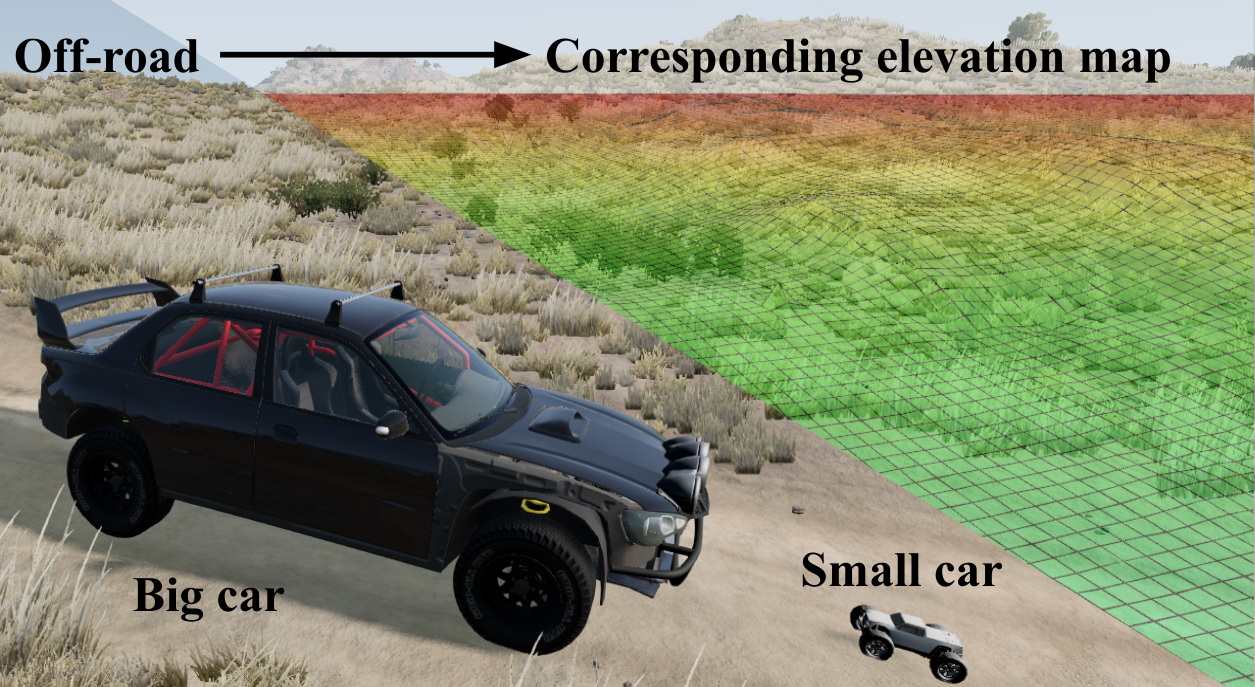}
  \caption{Elevation map replication for off-road map}
  \label{fig:offroad_sim}
\end{subfigure}%
\hspace{3pt}
\begin{subfigure}{.34\linewidth}
  \centering
  \includegraphics[width=\linewidth]{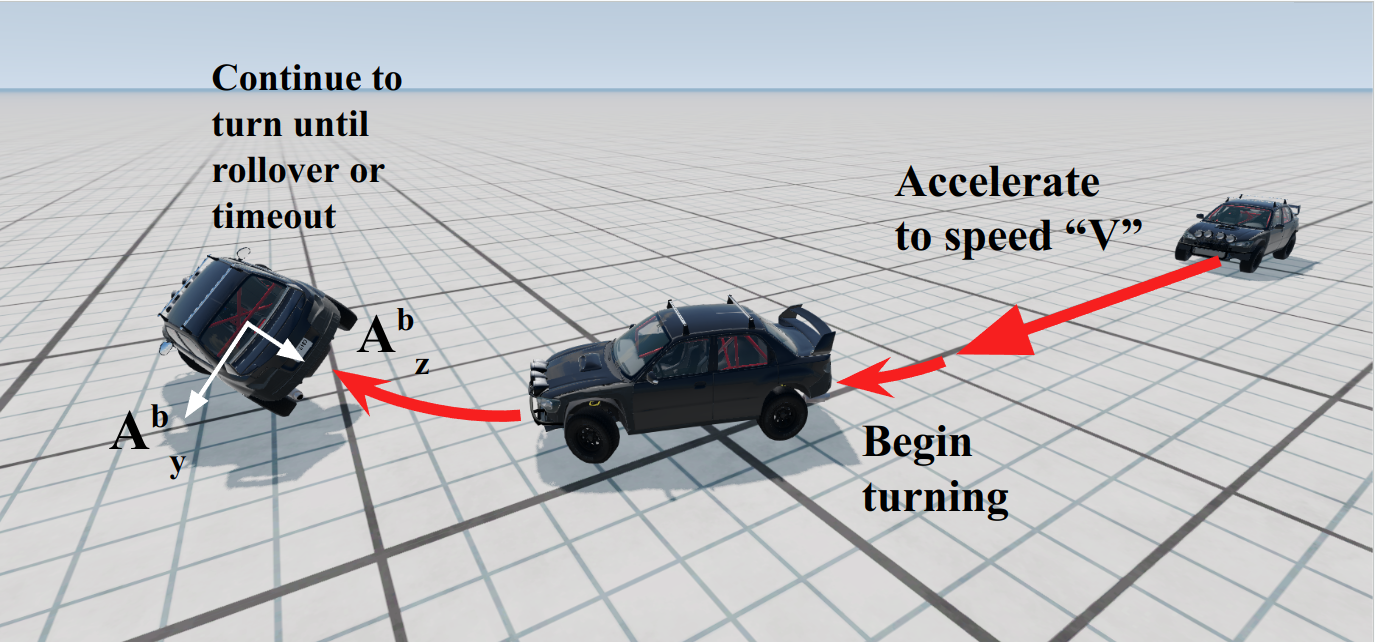}
  \caption{Isolated rollover setting on flat ground}
  \label{fig:isolated rollover}
\end{subfigure}
\vspace{-5pt}
\caption{
BeamNG's soft-body physics simulation (\ref{fig:beamNG_sim}) allows accurate simulation of second-order inertial effects, useful for problems such as rollover prevention(\ref{fig:isolated rollover}). 
The integration with BeamNG spoofs the elevation map(\ref{fig:offroad_sim}) for the off-road environment to isolate the control problem.
}
\vspace{-10pt}
\end{figure*}

\subsection{Rollover Prevention}\label{full_RPS}
While we build on existing work (see \ref{RPS_related}), we outline the specific implementation of rollover prevention used in our work as our constraints may differ from those of prior works.
The RPS is part of the Low-Level Controller. 
The Low-Level Controller allows some form of operator intervention at all times, thus, the operator can always prevent collisions based on visual contact. 
However, rollovers are caused by inertial effects not felt outside the vehicle, which makes them much harder for the operator to prevent.
The RPS is safety-critical, therefore, we avoid dependence on intrinsic and extrinsic state estimation, which may be erroneous. 
As such, the method outlined here only depends on wheel speed and IMU measurements. 
We also do not assume knowledge of the tire parameters and so, we assume no slip. 
We show that the dependence on this assumption is not strong.
Additionally, the RPS must not impose substantial computing requirements as it must run constantly in the background at a high update rate.
Due to these constraints on state estimation, knowledge of vehicle parameters, and compute power, model-based based approaches to rollover prevention, such as \citep{terrainCNN, terrainnet} do not address our problem.
We assume that the steering actuation is instantaneous.
Finally, we assume that the wheels remain in contact with the ground at all times. More details for the RPS and low-level controller can be found in Appendix \ref{LLC}

Consider a car turning on a sloped surface, which induces a positive roll angle $\phi$, being observed in a non-inertial frame.
The lateral and vertical accelerations ($A^b_y, A^b_z$  resp.) produce counteracting moments around the contact point of the right-side tires with arm lengths $H_{com}$, $L_t/2$, resp. where $H_{com}$ refers to the height of the center of mass above the contact surface under the vehicle and $L_t$ refers to the track width of the vehicle(see Fig. \ref{fig:coordinate_frame}). 
A rollover is imminent when the moment produced by $A^b_y$ exceeds that of $A^b_z$~\citep{force_angle_2000}.
Let this critical lateral acceleration be $A^c_y$.
\vspace{-3pt}
\begin{align}\label{force_angle}
\begin{split}
& |A^b_y| H_{com} < A^b_z L_t/2 \quad OR \quad |A^b_y| < A^c_y, \\
& A^c_y = A^b_z L_t/(2 H_{com}) 
\end{split}
\end{align}
We use the force-angle-measure as our rollover\citep{force_angle_2000} index(\textbf{RI}) which is the ratio of lateral to vertical acceleration.
\begin{align}\label{RI_def}
\begin{split}
& RI = A^b_y/A^b_z, \\
& RI_L = L_t/(2 H_{com})
\end{split}
\end{align}
Where $RI_L$ refers to the limiting or critical value of the rollover index beyond which a rollover is imminent($|RI| \geq RI_L$), following from the expression of $A^c_y$ in \ref{force_angle}.
Note that the term ``angle'' is used in the name as one can interpret the RI as $\arctan(A^b_y/A^b_z)$, though we use it as a ratio for mathematical convenience.

The RPS consists of two parts; the static limiter, and the feedback controller.
\subsubsection{Static Limiter}\label{static_limiter}
Assuming no slip, the wheel speed $V_w$ and steering input $\delta$ produce a centrifugal force that can be used to obtain the steering angle limit.
As the tires will always have some slip, this limiter will underestimate the steering angle limit, and so we also introduce a user-tunable ``slack'' $\delta_s$. We also account for gravity-induced lateral acceleration.
\vspace{-3pt}
\begin{align}\label{steering_limit}
\begin{split}
A^b_y &= V_w^2 \tan(\delta)/L_{fr}, \\
\delta_{C} &= \pm \tan^{-1}((A^c_y \mp g \sin({\phi})) L_{fr}/ V_w^2) \pm \delta_s \\
\end{split}
\end{align}
Where $L_{fr}$ represents the wheelbase.
Note that the slack variable allows for a configurable reduction in the conservatism that arises from the no-slip assumption and is tuned empirically on the platform.

\subsubsection{Feedback Control}\label{feedback_control}
Feedback control is necessary as the static limiter can suffer rollovers from model mismatch or environmental factors.
This also helps the end-user tune $\delta_s$; ideally, the feedback mechanism should never kick in, while using as much slack as possible.
The feedback controller uses IMU feedback to satisfy Eq. \ref{force_angle} with minimal steering angle change.
To handle non-linearities, we use feedback linearization before applying LQR to this system. 
Assume that for satisfying Eq. \ref{force_angle}, the small change in lateral acceleration is $\Delta A_y$, then (from \ref{steering_limit}) the required change in steering angle can be found from the lateral acceleration as:
\vspace{-3pt}
\begin{align}\label{acc_derivative}
\begin{split}
    \Delta \delta &= \Delta A_y (\cos^2(\delta) L_{fr}/V_w^2)
\end{split}
\end{align}
Observe that this approach would continue to adjust the steering angle until $\Delta A_y$ goes to 0, regardless of tire slip, which weakens the dependence on the ``no-slip'' assumption.
The LQR controller then provides a setpoint for the $\Delta A_y$ to regulate both the lateral acceleration error and the roll rate. 
The state transition equation, state penalty and control penalty matrices, assuming $\Delta t$ is the update time, are given by:
\begin{align}\label{LQR}
\begin{split}
&\begin{bmatrix}
    \frac{A^b_y}{A^b_z} - RI_L \\
    \omega_x \\
\end{bmatrix}_{t+1}
 = \begin{bmatrix}
1 & 0 \\
K & 1 \\
\end{bmatrix} 
\begin{bmatrix}
    \frac{A^b_y}{A^b_z} - RI_L \\
    \omega_x \\
\end{bmatrix}_{t}
+
\begin{bmatrix}
    1 \\
    K \\
\end{bmatrix} \Delta \frac{A^b_y}{A^b_z}, \\
&K = \Delta t A^b_z \frac{H_{com}}{J_x/m},\quad
Q = \begin{bmatrix}
10 & 0 \\
0 & 10 \\
\end{bmatrix},\quad
R = 1
\end{split}
\end{align}
Where $J_x, m$ refer to the roll moment of inertia and mass respectively. 
Note that in practice additional constraints may be used in Eq. \ref{force_angle} to reduce $A^c_y$ for safety reasons.

\subsection{Simulator interface}
We use BeamNG~\citep{beamng_tech}, a high-fidelity soft-body physics simulator geared towards mobile systems, that simulates physical systems as a collection of beams and nodes (see Fig. \ref{fig:beamNG_sim}). 
BeamNG has been used in the industry, as well as in academia~\citep{BeamNG1, BeamNG2}. 
We build our SITL and HITL stack around this simulator.
The SITL can replicate the elevation map (see Fig.~\ref{fig:offroad_sim}), useful for isolating control problems and benchmarking elevation mapping.
HITL is used for testing the software directly on the SBC (see Fig.~\ref{fig:architecture}).
For both the HITL and SITL, the simulator can run on a separate computer connected by WiFi/Ethernet, allowing accurate estimation of compute load on the SBC.
Note that our interface can be used for studying off-road autonomy in general, though we only use two particular vehicles for our experiments (see \ref{fig:offroad_sim}).

%% file: sections/Evaluation.tex
\begin{figure*}
\centering
\begin{subfigure}{.33\linewidth}
  \centering
  \includegraphics[width=\linewidth]{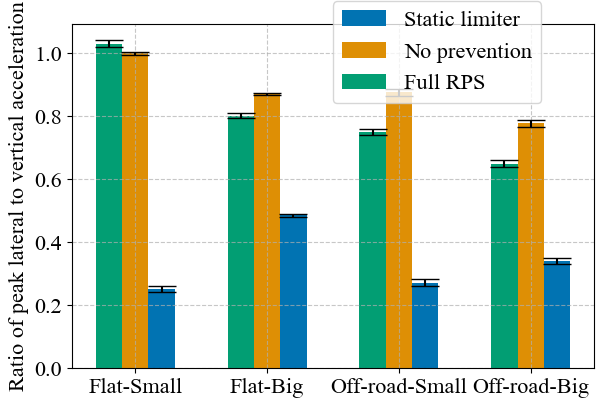}
  \caption{Peak $A^b_y/A^b_z$}
  \label{fig:lat_ratio}
\end{subfigure}%
\begin{subfigure}{.33\linewidth}
  \centering
  \includegraphics[width=\linewidth]{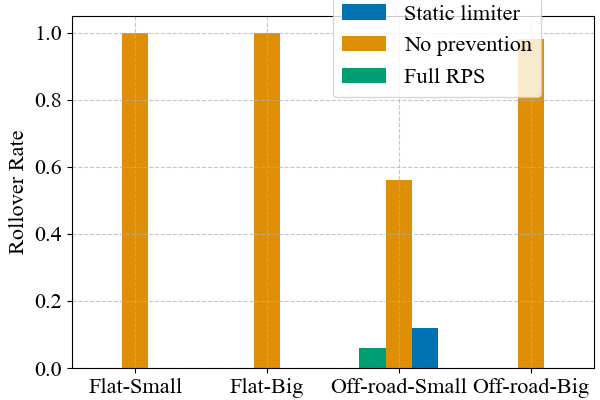}
  \caption{Rollover rate}
  \label{fig:rollover_rate}
\end{subfigure}
\begin{subfigure}{.181\linewidth}
  \centering
  \includegraphics[width=\linewidth]{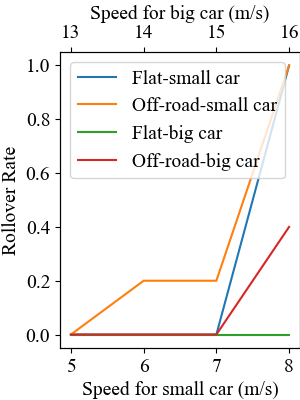}
  \caption{Rollover rate vs speeds}
  \label{fig:rollover_rate_speed}
\end{subfigure}
\begin{subfigure}{.119\linewidth}
  \centering
  \includegraphics[width=\linewidth]{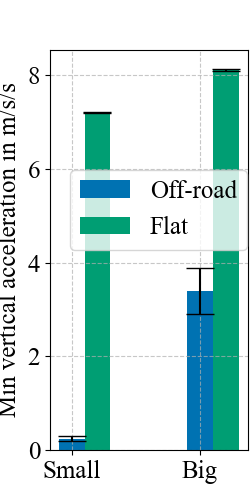}
  \caption{Min. $A^b_z$}
  \label{fig:rollover_vert_acc}
\end{subfigure}
\caption{
Static limiter obtains lower $A^b_y/A^b_z$ in \ref{fig:lat_ratio} and more rollovers(\ref{fig:rollover_rate}).
In 3 out of 4 scenarios, the rollover rate is 0 for both static limiter and full RPS in \ref{fig:rollover_rate}.
\ref{fig:rollover_rate_speed}, \ref{fig:rollover_vert_acc} show the breaking of instantaneous steering and constant ground contact assumption.
}
% \vspace{-15pt}
\label{isolated_exp}
\vspace{-5pt}
\end{figure*}

\section{Evaluation~}
\label{sec:evaluation}
\subsection{Experimental evaluation of simulator accuracy}
We evaluate how well the simulator predicts real-world behavior by comparing its dynamics prediction errors against the errors of two simpler mathematical models.
For simplicity, we perform this evaluation on a flat surface.

\textbf{Hypothesis H1}: The simulator's error is $<=$ error for either of the other models for all metrics.

\textbf{Scenario:}
In the real world, 3 trajectories, totaling $\approx100s$ of data, are collected on an outdoor, flat tiled surface at 50Hz.
Using IMU data, the coefficient of friction is estimated to be $\approx0.65\pm0.1$. Car specifications were measured directly.
The car is moved in circular patterns at speeds up to $\approx8m/s$, and accelerations up to $\approx7m/s^2$.
The dynamics model is initialized with real data only at the start of a new control sequence.

\textbf{Metrics:}
We calculate L2 norm errors for body frame acceleration, rotation rate, and velocity, normalizing them by the largest error among models to allow relative comparison.

\textbf{Models:}
\begin{itemize}
    \item BeamNG: represents the simulator's error.
    \item slip3d: represents the error for the model described in \ref{dynamics_model}.
    \item noslip3d: represents the error for a no-slip kinematic bicycle model~\citep{kinematic_flat}, with its velocity projected into 3D using the elevation map, as also done by~\citet{terrainnet}.
\end{itemize}

\begin{figure}
  \centering
  \includegraphics[width=\linewidth]{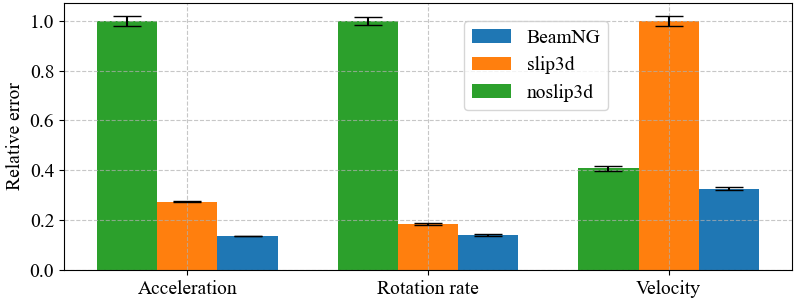}
  \caption{
  BeamNG performs as well as if not better than other models.
  }
  \label{fig:dynamics_comp}
  \vspace{-10pt}
\end{figure}

\textbf{Result:} Fig. \ref{fig:dynamics_comp} shows that the simulator's error is slightly lower than the next best in Velocity and rotation rate, and much lower in Acceleration($p<0.02$ for all). Note that all error bars represent 95\% confidence intervals. Thus, hypothesis \textbf{H1 is confirmed}.

\subsection{Validating RPS's utility as a safety system}\label{sim_exp}
We evaluate the utility of the RPS by evaluating how much it restricts the vehicle's maneuverability, and how much it prevents rollovers.
To stress the RPS, we set the tire friction for the small car to be 50\% more than the default and exaggerate the roll characteristics through suspension tuning.
Note that the simulator automatically reduces friction for off-road settings by $\approx20\%$. This, combined with the bumpiness of off-road terrain makes wheel-slip more likely in the off-road settings.
\begin{figure*}
\begin{subfigure}{0.47\linewidth}
  \centering
  \includegraphics[width=\linewidth]{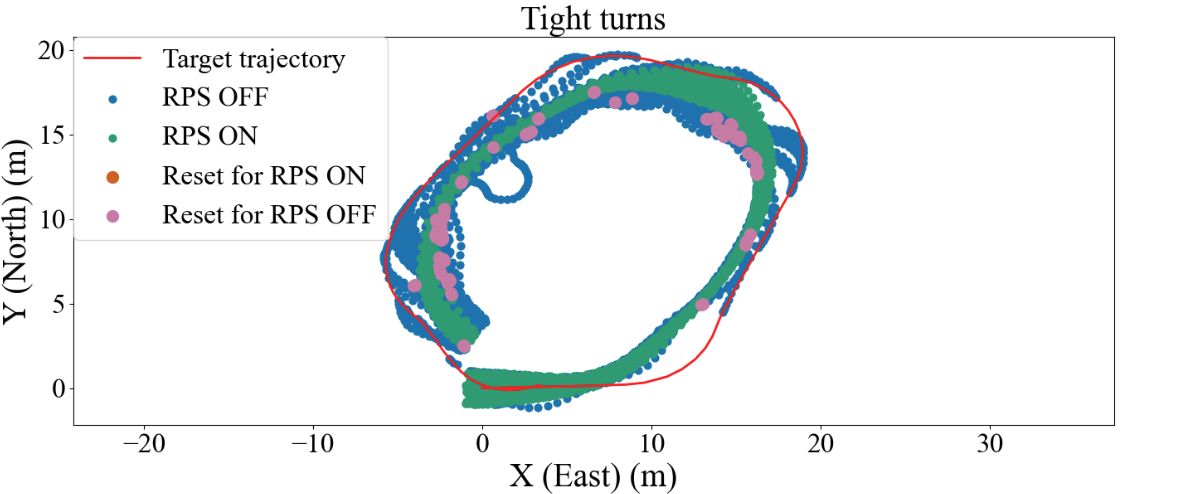}
  \caption{Trajectory plots for tight-turns test case}
  \label{fig:tight_turns}
\end{subfigure}%
\hspace{10pt}
\begin{subfigure}{0.47\linewidth}
  \centering
  \includegraphics[trim=0cm 0.14cm 0cm 0cm, clip, width=\linewidth]{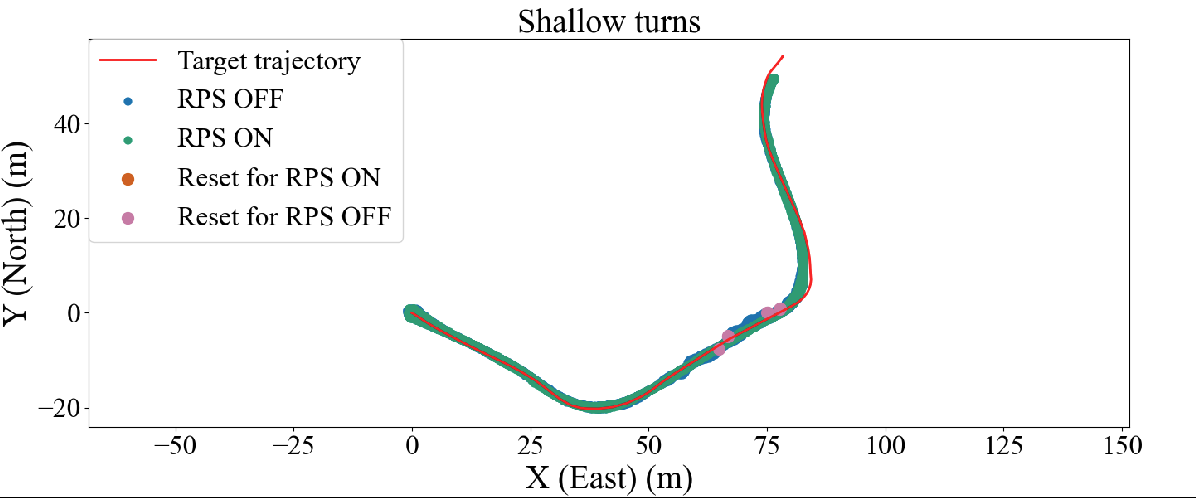}
  \caption{Trajectory plots for shallow-turns test case}
  \label{fig:shallow_turns}
\end{subfigure}%
\caption{
RPS reduces rollovers, not only when the vehicle needs to turn harder(Fig \ref{fig:tight_turns}) but also when the vehicle might rollover incidentally due to yaw instability from bumps in an off-road environment, indicated by the waviness of the RPS OFF trajectory and the high yaw accelerations $\alpha_z$ in Table \ref{in_the_loop_table}.
}
\label{closed_loop_exp}
\vspace{-5pt}
\end{figure*}
\begin{figure*}
\centering
\begin{subfigure}{0.33\linewidth}
  \centering
  \includegraphics[width=\linewidth]{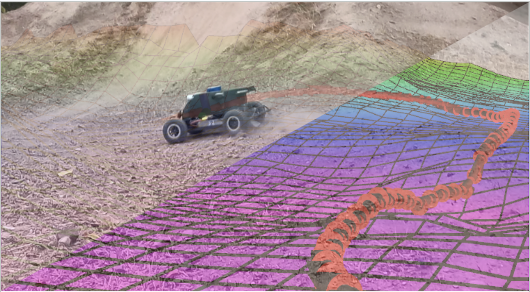}
  \caption{Off-road driving on uneven terrain}
  \label{fig:offroad_mpc}
\end{subfigure}
\begin{subfigure}{0.3\linewidth}
  \centering
  \includegraphics[width=\linewidth]{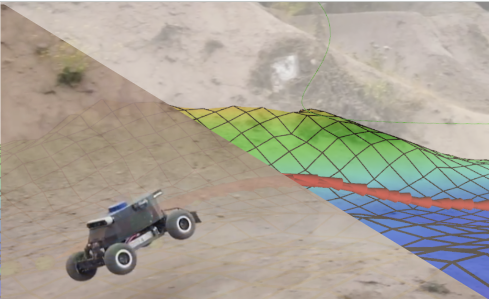}
  \caption{Airborne situations}
  \label{fig:offroad_jump}
\end{subfigure}%
\hspace{1pt}
\begin{subfigure}{0.32\linewidth}
  \centering
  \includegraphics[width=\linewidth]{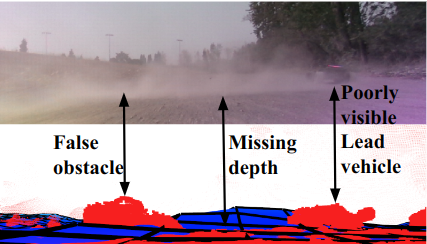}
  \caption{Sensor degradation, occlusion from dust}
  \label{fig:offroad_depth}
\end{subfigure}%
\caption{
High speed offroading \ref{fig:offroad_mpc}) often results in airborne situations(\ref{fig:offroad_jump}). Dust trails in multi-agent settings (\ref{fig:offroad_depth}) present interesting challenges.
}
\vspace{-10pt}
\label{real_world_data}
\end{figure*}
\subsubsection{Isolated evaluation of rollover prevention}\label{isolated_eval}
We test the RPS by forcing the vehicle into a rollover by applying maximum steering to one side while moving at a fixed speed $\textbf{V}$ (see Fig. \ref{fig:isolated rollover}).

\textbf{Hypothesis H2}:
The Full RPS(\ref{full_RPS}) achieves a greater ratio of peak $A^b_y/A^b_z$ than the static limiter(\ref{static_limiter}) while having the same or lower rollover rate.

\textbf{Scenarios:} 
We use the ``Small car'' and the ``Big car''(see Fig. \ref{fig:beamNG_sim}), over two terrains, Flat(Fig. \ref{fig:isolated rollover}) and Off-road(Fig. \ref{fig:offroad_sim}), for 50 iterations each. 
Fig. \ref{fig:isolated rollover} describes the setting, where \textbf{V} is increased linearly with each iteration, from $[4.8, 7.2] m/s$ for the small car, and $[9.6, 14.4] m/s$ for the big car. 
We also test the RPS's failure rate over a broader range of speeds, for 5 iterations at 4 speeds for each scenario.

\textbf{Metric:} We use the rollover rate and the peak ratio of lateral to vertical acceleration achieved once the car begins turning, up to a roll angle of 90 degrees.

\textbf{Algorithms:}
We compare ``No prevention'', ``Static limiter'' (\ref{static_limiter}) with $\delta_s$ set to 0 and ``Full RPS'' (\ref{full_RPS}) with $\delta_s$ set to 30 \% of the maximum steering angle. It should be noted that as platforms apart from ours shown in table \ref{comparison_table} do not posses RPS, they are represented by the ``No prevention'' algorithm in these experiments.

\textbf{Result:} Across all scenarios, full RPS achieves at least 83\% of the ratio achieved with no prevention.
Fig. \ref{fig:lat_ratio} and Fig \ref{fig:rollover_rate} show that the full RPS obtains a higher ratio of $A^b_y/A^b_z$ and a lower rollover rate when than the static limiter($p<0.02$ for all). Thus, hypothesis \textbf{H2 is confirmed}.

\subsubsection{In the loop evaluation of rollover prevention}\label{in_the_loop}
We test the RPS (\ref{full_RPS}) in conjunction with the high-level controller \textbf{(HLC)} (\ref{HLC}). 
To simulate dynamics mismatch, we set the friction coefficient used by the model to be 33\% lower than its value set in simulation.
\begin{table}[t!]
    \centering
    \setlength{\tabcolsep}{2.0pt} % Reduce space between columns
    \begin{tabular}{llllll}
    \toprule
        HLC with & Scenario & $TTC_{p}(s)$ & $TTC_{np}(s)$& $\alpha_z(rad/s^2)$ & $R.O._{avg}$\\
        \midrule
        RPS OFF & Tight & 11.04\std{1.56} & 8.76\std{1.47} & 207.74\std{157.77} & 2.28\\
        RPS ON & Tight & \textbf{8.39\std{0.13}} & \textbf{8.39\std{0.13}} & \textbf{170.60\std{18.45}} & \textbf{0}\\
        RPS OFF & Shallow & 17.63\std{0.76} & 17.47\std{0.59} & 210.05\std{86.92} & 0.16\\
        RPS ON & Shallow & \textbf{17.07\std{0.15}} & \textbf{17.07\std{0.15}} & \textbf{117.53\std{14.28}} & \textbf{0}\\
        \bottomrule
    \end{tabular}
    \caption{HLC w/ RPS ON performs at least as well as if not better than HLC w/ RPS OFF and makes the trajectories smoother, indicated by the lower $\alpha_z$}
    \label{in_the_loop_table}
\end{table}
\textbf{Scenarios:} 
We use the small car on flat ground, with two trajectories, for 50 iterations.
The first forces hard turning to reach the end, increasing rollover probability (Tight), and the second provides shallow turns and long straights (Shallow).
We randomly add noise to the start location for each iteration. 
Upon rollover, the vehicle is reset to the closest location on the target trajectory, and given a 1-second penalty to reflect the delay for resetting the vehicle in the real world. This is an optimistic comparison, as in the real world, rollovers can be fatal to the hardware.

\textbf{Hypothesis H3}:
The Full RPS system does not excessively interfere with the performance of the system.

\textbf{Metric:} We measure the time to reach the final goal with the 1-second penalty ($TTC_p$) and without it ($TTC_{np}$), maximum yaw acceleration just before rollover $\alpha_z$, and average number of rollovers ($R.O._{avg}$) for an iteration.

\textbf{Algorithms:}
We compare the HLC(\ref{HLC}) running without RPS (``RPS OFF'') against running it with RPS (``RPS ON'').
It should be noted that ``RPS OFF'' represents the approach of preventing rollovers by incorporating the rollover index into the cost function of a model predictive controller, as in \citep{terrainCNN, terrainnet}.

\textbf{Result:} Table \ref{in_the_loop_table}, shows that the completion time $TTC_p$ is lower with the RPS ON for the Tight turns scenario, and slightly lower for the Shallow turns scenario($p<0.02$ for both).
Thus, RPS does not interfere with HLC unless necessary, \textbf{confirming hypothesis H3}.
The $TTC_{np}$ in table \ref{in_the_loop_table} also shows that HLC with RPS ON is at least as good as if not slightly better even when not considering the 1-second penalty, from which we infer that using RPS does not jeopardize the performance of the system.

\subsection{Real world Evaluation}\label{real_world_eval}
\textbf{Scenario:} For real-world evaluation we only run the system with the RPS ``ON'', with $\delta_s = 0.3$, in the manual mode as well as the autonomous mode using the HLC(\ref{HLC}), using the default MPPI controller for waypoint following, with \textbf{no repairs} between runs.

\textbf{Hypothesis H4}:
The complete stack survives high lateral accelerations ($>8 m/s^2$) and speeds ($>5m/s$) over a large distance($>10km$) with minimal part damage.

\textbf{Metrics:} We measure the ``peak'' lateral acceleration $P(A^b_y)$, speed $P(V^b_x)$, delayed vertical acceleration just before rollover $D(A^b_z)$, number of rollovers \textbf{($R.O.$)}, and distance covered (S). 
The Peak values represent the top 99.7th percentile of the data, corresponding to $\approx 13$ seconds.
Note that the acceleration and speed values are low-pass filtered with a cutoff frequency of 2 Hz to remove large spikes in accelerations. 
An R.O. is when the roll angle $>1.0\mathrm{rad}$ while the vehicle speed $>0.5 m/s$. Delayed values are the values 0.2 seconds before the R.O.

\begin{table}[t!]
    \centering
    \setlength{\tabcolsep}{2.2pt} % Reduce space between columns
    \begin{tabular}{rlllll}
    \toprule
        Mode & $P(A^b_y)(m/s^2)$ & $P(V^b_x)(m/s)$& $R.O.$ & $D(A^b_z)(m/s^2)$ & S(Km) \\
        \midrule
        Auto & 7.41 & 6.95 & 2 & 3.29 & $\approx$12\\
        Manual & 8.92 & 8.01 & 1 & 0.75 & $\approx$38\\
        \bottomrule
    \end{tabular}
    \caption{Low $D(A^b_z)$ indicates weightlessness or loss of ground contact}
    \label{perf_table}
\vspace{-10pt}
\end{table}

\textbf{Result:} Real world experiments over $\approx$50Km (Table \ref{perf_table}) show that the complete stack obtains speeds $>5 m/s$, lateral accelerations above $>8 m/s^2$, and 3 rollovers (R.O.).
The rollovers damage only a sacrificial, easy-to-replace LiDAR mount in the last run (see Appendix \ref{lidar_appendix}). Over the 50 kilometers, the mean and standard deviation of the speed was $(2.18, 2.38) m/s$, and the mean and standard deviation of the lateral acceleration was $(2.49, 2.57) m/s^2$.
Thus, hypothesis \textbf{H4 is confirmed}. For further qualitative assessment of our real-world experiments, we refer the reader to the publicly released dataset.
\vspace{-10pt}

%% file: sections/Discussion.tex
\section{Discussion~}
\balance
\label{sec:discussion}
\textbf{Autonomy stack:}
The autonomy stack presented in this work provides a useful reference point that combines relatively mature state-of-the-practice methods.
Considering the low inertia of the system, we aimed for a sub-50 ms perception-to-control latency in the autonomy stack(see Fig.\ref{fig:architecture}). 
Due to SWaP constraints, this precluded using additional capabilities, such as semantic segmentation~\citep{realtime_seg}, as well as the use of learned dynamics models~\citep{terrainCNN} without increasing the latency or memory requirements. We believe, however, that a standardized platform such as this can make it easier for researchers to improve individual sub-components over time, such that additional capabilities become available eventually.

\textbf{Choice of simulator and simulation fidelity:}
While many good simulators for robotics exist today, we chose BeamNG as it is geared specifically towards simulating vehicles, and is used by the industry for the same.
It is geared towards simulating vehicle damage, which is particularly useful for inferring whether an off-road autonomy stack is safe or not, without having to create hand-crafted constraints on velocities, and accelerations, which can create misaligned incentives if the simulator is being used for training a reinforcement learning agent.
We do not preclude future extensions that allow the use of other simulators.

We only verify simulation fidelity for operation on flat surfaces. Replication of real-world elevation maps, their integration into the simulator, and the acquisition of precise GPS coordinates to obtain more accurate benchmarks can be addressed by future work.

\textbf{Reducing cost and increasing accessibility}
In Table \ref{comparison_table}, we compare the cost of a 1/5th scale platform (AutoRally~\citep{goldfainthesis}) with our system that uses a 1/10th scale platform. While a bigger vehicle would be more expensive, note that the cost of sensing and computing is not as closely tied to the size, and is higher due to the use of industrial-grade components.

We find from 50 kilometers of real-world testing that low-cost commercial-grade hardware can indeed be used for such a platform. 
We believe that this is at least in part made possible through the use of the RPS safety mechanism. While a few rollovers would not cause significant damage to our platform, a rollover every 100 meters of travel would incur 500 rollovers for 50 kilometers, requiring repeated repairs, and incurring a greater cost over time than using ruggedized hardware.

\textbf{Notes on rollover prevention:}
It should be noted that in the \ref{in_the_loop} experiment, while the high-level controller could use either adaptive control methods or online sys-ID as in ~\citep{dust_mpc} to address a dynamics mismatch, it is introduced on purpose to demonstrate how the RPS can help in the case of a faulty high-level controller.
From Table \ref{in_the_loop_table}, it can be seen that the dynamics mismatch causes the HLC to apply larger than necessary inputs that result in higher yaw acceleration($p<0.02$).
In contrast, corrections from RPS smoothen out the trajectories (see Fig. \ref{closed_loop_exp}).
In contrast to the isolated setting(see Fig. \ref{fig:rollover_rate_speed}), in the closed loop experiments there are no rollovers (see table \ref{in_the_loop_table}) despite the peak speeds exceeding $8.0$ m/s for both scenarios, as the HLC does not provide unreasonable commands to cause rollovers on purpose.

In our isolated experiments, we also push the system to unreasonably high speeds to understand the limits of the safety mechanism. 
The unevenness of off-road surfaces results in airborne (Fig. \ref{fig:offroad_jump}) situations, apparent from the lower vertical acceleration(Fig. \ref{fig:rollover_vert_acc}), breaking the assumption of constant ground contact -- broken more often in the real world(Table \ref{perf_table}, Fig. \ref{fig:offroad_jump}).
To the best of our knowledge, autonomous mid-air control of cars is currently an open problem that could be addressed by future work.

In Fig\ref{fig:lat_ratio}, it can also be seen that the lateral acceleration ratio for the small car on the flat surface is slightly ($\approx3\%$) higher when using RPS as compared to when not using any prevention. 
Note that due to the vehicle's suspension, the body of the vehicle can roll slightly even before $RI_L$ is reached. 
Due to an effect known as ``jacking''\citep{jacking} the center of mass moves upwards relative to the outside tire's contact point due to this body roll.
Once the center of mass moves upwards, the $RI_L$(see Eq. \ref{RI_def}) becomes smaller, creating a positive feedback loop; requiring less lateral traction to sustain the rollover.
This happens much faster on the smaller vehicle owing to its lower roll inertia ($\approx3.5$ times faster on average).
When a rollover is prevented, the center of mass is kept low.
This allows the system to sustain slightly higher lateral accelerations.
Note that we consider this to be an unintended effect of the rollover prevention system -- not something that the system was designed to do.

\textbf{Sensor degradation:}
During our field experiments, we observed that in dry off-road environments, apart from sensor degradation from dust coverage~\citep{Off_road_foundation}, dust trails from a lead vehicle can cause issues such as false depth or no depth (see \ref{fig:offroad_depth}.
Depending on the density of the dust particles, this may obstruct lidar as well, which can create interesting situations for multi-agent off-road navigation problems that future work could address.

\textbf{Licensing: }
The HOUND software stack is built using multiple open-source components, with either MIT, BSD-3, or GPL-v3 licenses. The BeamNG simulator is an exception to this. While the HOUND-BeamNG integration is open-source, the simulator itself is not. However, at the time of writing, BeamNG is available for academic purposes at no cost.

%% file: sections/Conclusion.tex
\section{Conclusions and Future Work} \label{sec:conclusion}
The paper presents an inexpensive, 1/10th scale research platform for \textbf{H}igh-speed \textbf{O}ff-road \textbf{U}nder-actuated \textbf{N}on-holonomic \textbf{D}riving(\textbf{H.O.U.N.D.}).
HOUND integrates state-of-the-art methods for terrain mapping, localization, and model predictive control geared towards aggressive offroad driving into a single platform.
The platform is integrated with a high-fidelity vehicle simulator, BeamNG, to allow both software and hardware-in-the-loop testing.
We deploy HOUND in the real world, at high speeds, on four different terrains covering 50 km of driving, and highlight the utility of rollover prevention for traversing difficult terrain at high speed through real-world experiments.

At the moment, the perception system uses elevation maps to represent the environment.
This requires less computation than a 3D representation and is convenient for predicting vehicle dynamics.
However, the assumptions imposed on the perception and dynamics models make them a lossy compression of the real world.
For instance, an elevation map might consider a fallen log of wood as a hill, leading to errors in physics prediction, which may lead to errors in navigation.
Future work may explore how this problem may be addressed in a compute and memory-efficient manner for real-time operation on such robots.
Smaller robots are safer and easier to operate, which makes dataset generation easier compared to using full-size vehicles.
Future work should investigate the re-targeting of datasets collected on small platforms to their use on large and full-size platforms.

%% file: sections/Acknowledgments.tex
\section*{Acknowledgements}
\small{
This work was (partially) funded by the National Science Foundation NRI (\#2132848) and CHS (\#2007011), DARPA RACER (\#HR0011-21-C-0171), the Office of Naval Research (\#N00014-17-1-2617-P00004 and \#2022-016-01 UW), and Amazon. We thank Alex Lin for support.
Authors are with the Paul G. Allen School of Computer Science \& Engineering, University of Washington, Seattle, WA, USA. Email: {\tt\small \{sidtalia, schmttle, cmavro, spitzer, lambert6, siddh \}@cs.washington.edu}.
\normalsize

%% file: sections/Appendix.tex
\vspace{-10pt}
\section{Appendix~}\label{appendix}
\begin{figure}[!htb]
\centering
\includegraphics[width=\linewidth]{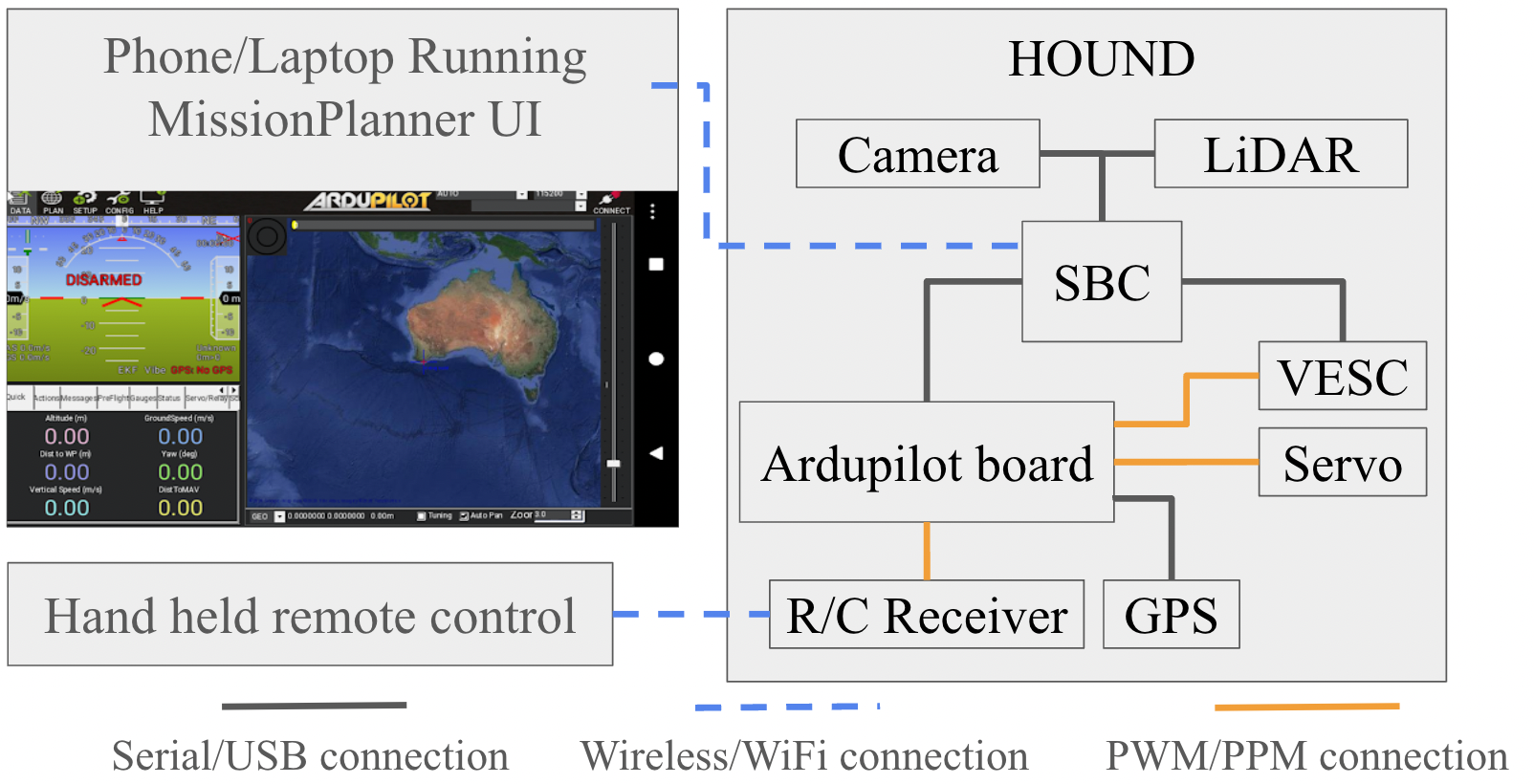}
\caption{Illustration describing how the components are connected to each other}
\label{fig:connection diagram}
\end{figure}
\vspace{-15pt}

\subsection{Utilising the Ardupilot ecosystem:} \label{ardupilot_appendix}
The Ardupilot ecosystem provides many plug-and-play features that we take advantage of in this work.
It takes care of time synchronization between itself and the SBC, as well as provides certain safety features for field operations.

As the Ardupilot board acts as a bridge between the SBC and the actuators, it allows the operator to always be in control of the final commands being sent to the hardware.
For instance, the ardupilot board can ``disarm" the vehicle under different circumstances by setting the motor command to ``0", preventing the car from moving of its own volition or hitting the brakes if it is currently moving.
The board can be, and by default is, configured to disarm the vehicle either through a two-position switch on the hand-held remote, or by the loss of radio contact with said remote.
It can additionally be configured to disarm if the vehicle moves outside of a particular region, generally known as ``geofencing"~\citep{geofencing}, or if the vehicle loses the GPS signal completely during autonomous operation. 
Note that in the case of GPS loss, operation in manual modes is still possible.

When an internet connection is available to the SBC, RTCM~\citep{RTK} corrections can be obtained using NTRIP~\citep{NTRIP}, which can be forwarded to the GPS by the ardupilot board to obtain high-accuracy RTK positioning.
During field operations, the MissionPlanner~\citep{mission_planner} UI running either on the user's laptop or phone (see Fig \ref{fig:connection diagram}), connected to the SBC over WiFi, can be used to calibrate the IMU and compass in the ardupilot board, change configuration parameters, or reboot it.
The UI also provides a convenient way of sending GPS waypoints to the SBC, which can then be used by the high-level controller, or a local planner.
Additionally, the onboard buzzer on the ardupilot boards is used for producing notification sounds that let the user know the internal status of the sub-systems. For instance, the notification sounds can let the user know if the perception system has started running yet or not, or if the system has started or stopped recording data. If the user only wishes to collect data from the platform with manual driving, the perception stack and low-level controller can be configured to auto-boot, and the user can perform this task without needing to carry a laptop with them to monitor the vehicle's internal status.

\begin{figure*}[!htb]
\centering
\begin{subfigure}{.33\linewidth}
  \centering
  \includegraphics[width=\linewidth]{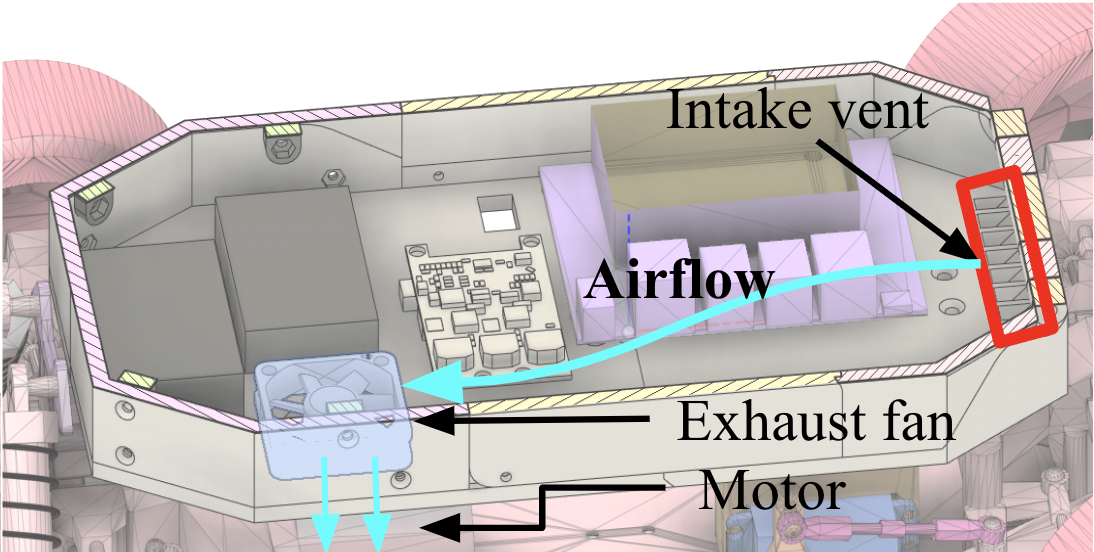}
  \caption{Heat management using negative pressure}
  \label{fig:heat_mgmt}
\end{subfigure}%
\hspace{1pt}
\begin{subfigure}{.33\linewidth}
  \centering
  \includegraphics[width=\linewidth]{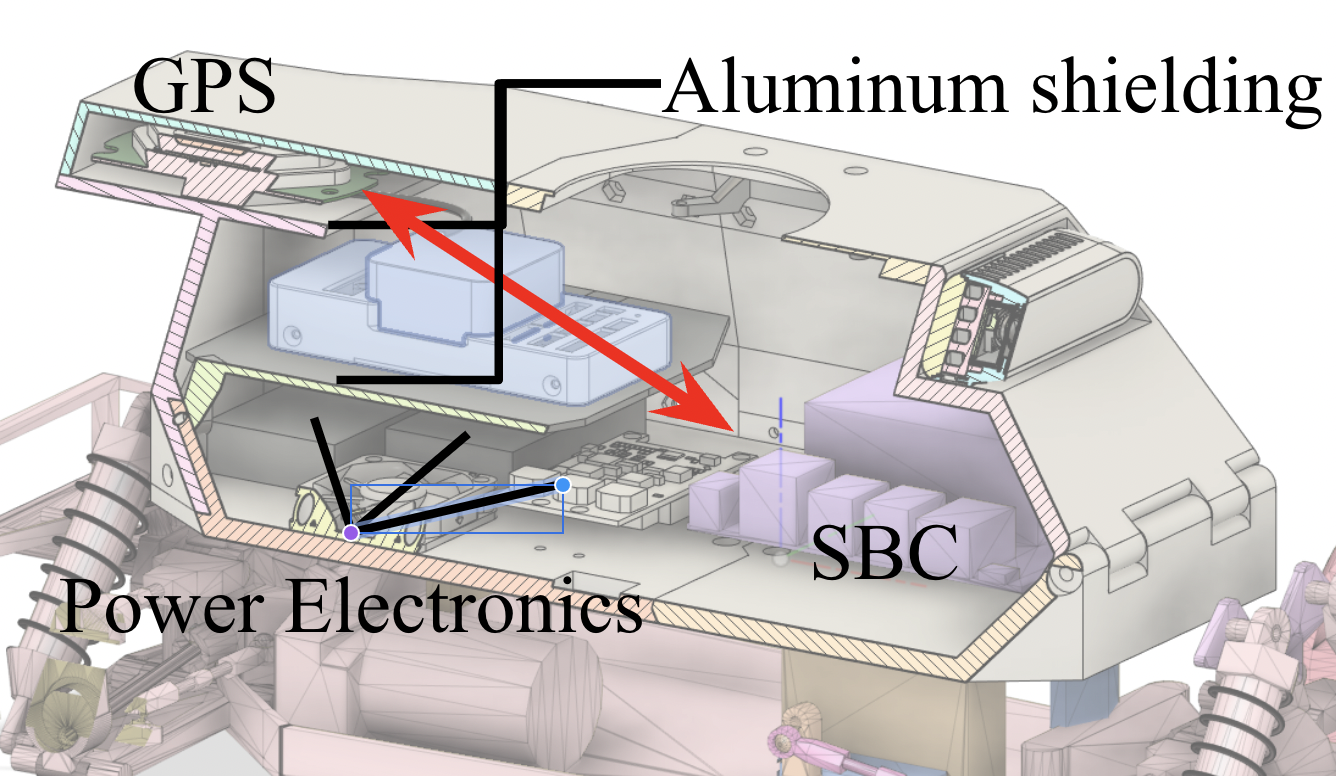}
  \caption{Managing electromagnetic interference.}
  \label{fig:emi_mgmt}
\end{subfigure}
\begin{subfigure}{.28\linewidth}
  \centering
  \includegraphics[width=\linewidth]{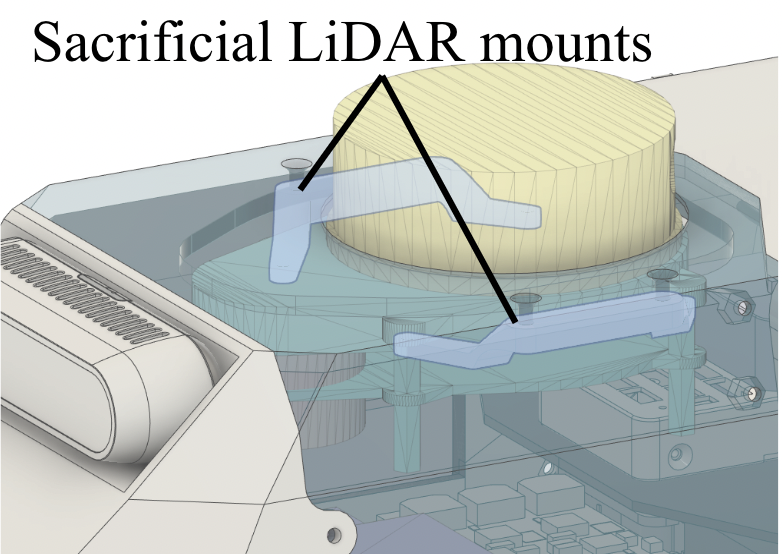}
  \caption{Sacrificial LiDAR mounts}
  \label{fig:lidar_mounts}
\end{subfigure}
\caption{
Small enclosures create challenges for heat management \ref{fig:heat_mgmt} and electromagnetic interference \ref{fig:emi_mgmt}. Sacrificial LiDAR mounts absorb the impact energy during incidental rollovers to reduce damage to the LiDAR.
}
\label{hardware_design}
\vspace{-15pt}
\end{figure*}

\subsection{Hardware design decisions}\label{hardware_appendix}
\textbf{Heat management: }
To protect the system from environmental factors as well as collisions, most of the hardware is placed in an enclosed shell (see Fig \ref{fig:coordinate_frame}).
We use an active negative pressure cooling system, where the fan creates a slight negative pressure in the shell, and routes the air around all the components before exhaustion (see Fig \ref{fig:heat_mgmt}).
The exhaust goes over the motor, extracting the motor's heat as well. While the exhaust air is hotter than the ambient, it is still much cooler than the motor's temperature and can provide the necessary airflow for sustained cooling.
The intake vent is positioned and sized to minimize the entry of dust and dirt.
While the default setup uses negative air pressure, a positive pressure system would work as well, requiring a dust filter under the fan and flipping the orientation of the fan.

\textbf{Electromagnetic interference management: }
The SBC, devices that use USB-3.0 protocol, and power converters, all produce electromagnetic interference(\textbf{EMI}) that affects the GPS's accuracy~\citep{usb_interference}.
On big platforms, the spacing between the GPS and such EMI-generating devices can be large, and so this may or may not be a concern for them.
On the HOUND, we minimize the impact of the EMI on the GPS by one, maximizing the separation between the SBC and the GPS, and two, by placing two layers of Aluminum shielding between it and the power-electronics (see Fig \ref{fig:emi_mgmt}). The Aluminum shielding is created using Aluminum tape, readily available from any hardware store, or online. 

\textbf{Sacrifical LiDAR mounts: }\label{lidar_appendix}
As we show in our experiments, rollover prevention is not without limitations. Eventually, the vehicle is bound to roll over. 
The LiDAR becomes the most susceptible sensor during a rollover as it protrudes significantly out of the body shell.
If the LiDAR were to be mounted directly onto the roof, one of two things could happen, either the LiDAR survives but the roof caves in, or the roof survives but the LiDAR is damaged. The mid-section, on which the LiDAR is mounted, can take at least 10 hours to print, and the LiDAR costs at least \$100. An alternate design choice would be to create a ``roll cage" around the lidar, however, that would now obstruct the LiDAR's view.
To avoid damage to either we use sacrificial mounts to attach the LiDAR to the roof, designed to be strong enough to survive low-speed rollovers but to absorb energy and break on high-speed impacts (see Fig \ref{fig:lidar_mounts}).
These parts can be printed much faster and can reduce damage to the LiDAR.

\subsection{Low Level Controller}\label{LLC}
The Low-Level Controller is responsible for sending the motor control and steering control commands to the Ardupilot board, where they are converted to appropriate pulse-width-modulated signals.
It operates in two modes -- manual and autonomous, where the RPS is ``ON" by default in both.
In the manual mode, the low-level controller listens to control commands from the user's hand-held remote control (\ref{fig:connection diagram}). 
In the autonomous mode, it listens to the control commands from the high-level controller, however, the user's throttle input corresponds to the wheel speed limit.
For instance, if the operator visually infers that a collision is about to occur, and the vehicle is in autonomous mode, the user can reduce the speed limit to prevent the collision. 
Alternatively, the user can slowly increase the speed limit as they gain confidence in their custom controller.
This speed limit is also published on a separate ROS topic should the high-level controller need it.
For instance, in our implementation of the MPPI, we change the sampling range according to the speed limit imposed by the user, such that wheel speeds above this limit are not sampled.

\textbf{wheelspeed control: } 
The wheel speed control takes as input the wheel speed target $V^*_w$ and produces a duty cycle signal for the motor using a closed-loop PI controller with a feed-forward term. Here, the current wheel speed $V_w$ is obtained from the VESC.
\begin{equation}
    D = \frac{E_{i}}{E_{n}} K_f V^*_w + K_p (V^*_w - V_w) + K_i \int_0^t (V^*_w - V_w) \, d\tau
\end{equation}
Where $D$ refers to the motor duty cycle, $E_{i}, E_{n}$ represent the instantaneous and nominal battery voltage, $K_f, K_p, K_i$ refer to the feedforward gain, proportional gain, and integral gain.
The feed-forward term's gain is obtained by measuring the ratio of the duty cycle to wheel speed on a flat tarmac surface for a small range of speeds, and the gains for the proportional and integral terms are tuned such that changes in slope or surface can be dealt with in real-time.

\textbf{Note on wheel speed measurement: }
To simplify the sensing apparatus, we assume that all the wheels rotate at the same wheel speed. 
We find that this assumption works well for high-speed driving ($> 2 m/s$) on uneven, smooth terrain, but begins to break down when dealing with extremely rough terrain, such as in rock-crawling, due to the use of open differentials in the vehicle's drivetrain~\citep{dynamic_gerdes}. The work by ~\citet{vertiwheeler} addresses this by having separate measurements for each wheel, however, it comes at the cost of additional build complexity.

\textbf{Why the RPS only adjusts steering angle:}
In \ref{full_RPS}, we only adjust the steering angle for two reasons.
First, the rate at which lateral acceleration can be changed (lateral jerk) by the steering is much higher than what is achieved by changing the vehicle's velocity.
Consider the following example; when turning at a speed of $3.0m/s$ with a friction coefficient of $1.0$, the steering angle would be $\approx 15$ degrees. The steering can move from $+15$ to $-15$ degrees in 0.1 seconds, based on the servo specifications. This results in a lateral jerk of $\approx 200 m/s^3$.
On the other hand, when turning, only a small fraction of the grip is available for changing speed.
For the sake of the argument, assume that this is not the case, and all of $9.81 m/s^2$ of longitudinal acceleration is available. 
Within 0.1 seconds, the effective lateral jerk would still be $\approx 50m/s^3$, which is 4x lower than what would be obtained from the steering.

Second, slowing down while turning can momentarily increase the lateral acceleration due to a phenomenon generally known as ``lift-off oversteer"~\citep{liftoff_oversteer}.
Intuitively, reducing the speed --done to prevent a rollover due to excess lateral acceleration-- moves more weight onto the front tires almost instantly.
This momentarily reduces the turning radius, without significantly reducing the vehicle's velocity.
This increase in lateral acceleration then results in a rollover.
The exact extent to which this happens, and therefore preventing it, would require knowledge of the vehicle's body frame velocity as well as the tire parameters, which we chose not to depend on.